\documentclass[journal]{IEEEtran}
\ifCLASSINFOpdf
\else
\fi
\usepackage{amsmath} 
\usepackage{amsfonts} 
\usepackage{graphicx}
 \usepackage{color}
\usepackage[lofdepth,lotdepth]{subfig}

\usepackage{algorithm}
\usepackage{algorithmic}

\newcommand{\Reals}{\mathbb{R}} 
\newcommand{\fracpartial}[2]{\frac{\partial {#1}}{\partial {#2}}}
\newcommand{\fpevalat}[3]{\left( \frac{\partial {#1}}{\partial {#2}} \right)_{#3}}
 
\newcommand{\vecr}[0]{\vec{r}}
\newcommand{\vecx}[0]{\vec{x}}
\newcommand{\vecv}[0]{\vec{v}}
\newcommand{\vecpInBPTTalg}[0]{\vec{p}}
\newcommand{\planeConstantP}[0]{\vec{P}}
\newcommand{\vecn}[0]{\vec{n}}
\newcommand{\veca}[0]{\vec{u}}
\newcommand{\Weightsz}{\vec{z}}
\newcommand{\Gapprox}{{\widetilde{G}}}
\newcommand{\GapproxFull}[1]{{\Gapprox(\vecx_{#1},\Weights)}}
\newcommand{\Vapprox}{{\widetilde{V}}}
\newcommand{\VapproxFull}[1]{{\Vapprox(\vecx_{#1},\Weights)}}

\newcommand{\model}[0]{f}
\newcommand{\modelFull}[1]{\model(\vecx_{#1}, \veca_{#1})}
\newcommand{\reward}[0]{U}
\newcommand{\rewardFull}[1]{\reward(\vecx_{#1}, \veca_{#1})}
\newcommand{\modelClipped}[0]{{f^C}}
\newcommand{\modelClippedFull}[1]{\modelClipped(\vecx_{#1}, \veca_{#1}, \planeConstantP, \vecn)}
\newcommand{\rewardClipped}[0]{{U^C}}
\newcommand{\rewardClippedFull}[1]{\rewardClipped(\vecx_{#1}, \veca_{#1}, \planeConstantP, \vecn)}
\newcommand{\rewardFinal}[0]{\Phi}
\newcommand{\rewardFinalFull}[1]{\rewardFinal(\vecx_{#1})}

\newcommand{\StateSpace}[0]{\mathbb{S}}
\newcommand{\TerminalStateSet}[0]{\mathbb{T}}
\newcommand{\Rpi}[0]{J}
\newcommand{\RpiFull}[1]{\Rpi(\vecx_{#1},\Weightsz)}
\newcommand{\RpiClipped}[0]{J^C}
\newcommand{\RpiClippedFull}[1]{\RpiClipped(\vecx_{#1},\Weightsz)}
\newcommand{\dRdx}[0]{\fracpartial{\Rpi}{\vecx}}
\newcommand{\dRdxEvalAt}[1]{\left(\dRdx\right)_{#1}}
\newcommand{\Weights}{\vec{w}}
\newcommand{\PolicyFunction}{A}
\newcommand{\PolicyFunctionFull}[1]{\PolicyFunction(\vecx_{#1}, \Weightsz)}

\newcommand{\actiona}[0]{u}

\newcommand{\ifdf}[1]{#1}
\newcommand{\df}[0]{\ifdf{\gamma}}
\newcommand{\dfa}[1]{\ifdf{\gamma{#1}}}
\newcommand{\half}[0]{{\frac{1}{2}}}
\newcommand{\dJdz}[0]{\fracpartial{\Rpi}{\Weightsz}}
\newcommand{\dJdw}[0]{\fracpartial{\Rpi}{\Weightsz}}

\newcommand{\lineParameter}[0]{\lambda}
\newcommand{\clippingFraction}[0]{\Lambda}
\newcommand{\isDefinedToBe}[0]{:=}
\newcommand{\clippingFractionFull}[1]{\clippingFraction(\vecx_{#1}, \veca_{#1}, \planeConstantP, \vecn)}
\newcommand{\vecDelta}{\vec{e}}

\newcommand{\hdpTarget}{V_{t+1}}
\newcommand{\dQdx}{{Q_x}}
\newcommand{\dQdu}{{Q_u}}
\newcommand{\finalTimestep}{{T}}
\newcommand{\cartPoleForce}{F}
\newcommand{\expectation}[1]{\left<{#1}\right>}
\newcommand{\learningRateCritic}{\beta}
\newcommand{\learningRateActor}{\alpha}
\begin{document}
\graphicspath{{./}{diagrams/}{resultsLL/}{resultsCP/}{resultsWilliamsBPTT/}{results/}}
%
\title{The Importance of Clipping in Neurocontrol by Direct Gradient Descent on the Cost-to-Go Function and in Adaptive Dynamic Programming}

\author{Michael~Fairbank
\thanks{M. Fairbank is with the Department of Computing,
School of Informatics,
City University London, 
London, UK.  email: michael.fairbank@virgin.net}
}

\maketitle

\begin{abstract}
In adaptive dynamic programming, neurocontrol and reinforcement learning, the objective is for an agent to learn to choose actions so as to minimise a total cost function.   In this paper we show that when discretized time is used to model the motion of the agent, it can be very important to do ``clipping'' on the motion of the agent in the final time step of the trajectory.  By clipping we mean that the final time step of the trajectory is to be truncated such that the agent stops exactly at the first terminal state reached, and no distance further.  We demonstrate that when clipping is omitted, learning performance can fail to reach the optimum; and when clipping is done properly, learning performance can improve significantly.  

The clipping problem we describe affects algorithms which use explicit derivatives of the model functions of the environment to calculate a learning gradient.  These include Backpropagation Through Time for Control, and methods based on Dual Heuristic Dynamic Programming.  However the clipping problem does not significantly affect methods based on Heuristic Dynamic Programming, Temporal Differences or Policy Gradient Learning algorithms. Similarly, the clipping problem does not affect fixed-length finite-horizon problems.

\end{abstract}

\IEEEpeerreviewmaketitle

\section{Introduction} \label{sec:introduction}

\begin{figure}[h]
\centering
\scalebox{0.9}{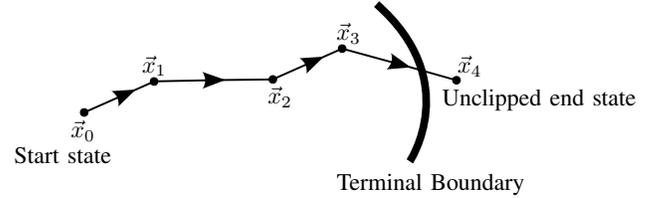}
\caption{A trajectory reaching a terminal state.  The thick curved line indicates a boundary of terminal states.  In this diagram, clipping does not take place, and the trajectory penetrates beyond the terminal boundary.  When clipping is used correctly, we intend to stop the agent exactly at the point of intersection between the trajectory and the terminal boundary.}
\label{fig:hittingTerminalBoundary}
\end{figure}

In Adaptive Dynamic Programming (ADP) \cite{adpAnIntroduction}, Neurocontrol \cite{Werbos92neurocontrol}, and Reinforcement Learning (RL) \cite{suttonbarto-1998}, an agent moves in a state space $\StateSpace \subset \Reals^n$, such that at integer time step $t$, it has state vector $\vecx_t\in \StateSpace$.  $\TerminalStateSet$ is a fixed set of {\it terminal states}, with $\TerminalStateSet \subset \StateSpace$.  At each time $t$, the agent chooses an action $\veca_t$ which takes it to the next state according to the environment's model function \begin{align}
\vecx_{t+1} = \modelFull{t}, \label{eqn:stateTransitionFunction}
\end{align}
thus the agent passes through a trajectory of states $(\vecx_0, \vecx_1,\vecx_2, \ldots)$, terminating only when (and if) a terminal state is reached, as illustrated in Fig. \ref{fig:hittingTerminalBoundary}.     As shown in this figure, clipping is the concept of calculating the exact fraction in the final time step at which a boundary of terminal states is reached, and stopping the agent exactly at this boundary.  The name clipping is taken by analogy to the concept in computer graphics.  Without clipping, the discretization of time would cause the agent to  penetrate slightly beyond the terminal boundary, as shown in the figure.

On transitioning from each state $\vecx_t$ to the next, the agent receives an immediate scalar cost $\reward_t$ from the environment according to the function \begin{align}
\reward_t \isDefinedToBe \rewardFull{t}. \label{eqn:instanteousReward}                                                                                                                                                               \end{align}
In addition, if the agent reaches a terminal state $\vecx \in \TerminalStateSet$, then an additional terminal cost is given by the scalar function $\rewardFinalFull{}$.  

Throughout this paper, subscripts on variables will be used to indicate the time step of a trajectory.  And from now on in the paper, we will only consider episodic, or finite horizon, environments; that is environments where all trajectories are guaranteed to meet a terminal state eventually.

The ADP problem is for the agent to learn to choose actions so as to minimise the expectation of the total long-term cost received from any given start state $\vecx_0$.  Specifically, the problem is to find an {\it action network} $\PolicyFunctionFull{}$, where $\Weightsz$ is the parameter vector of a function approximator, which calculates an action \begin{align}
\veca_t=\PolicyFunctionFull{t} \label{eqn:policyFunction}\end{align}
to take for any given state $\vecx_t$, such that the following long-term cost is minimised:
\begin{equation}
\RpiFull{0}\isDefinedToBe \expectation{\sum_{t=0}^{\finalTimestep-1} \dfa{^t} \reward_{t}+\dfa{^\finalTimestep}\rewardFinalFull{\finalTimestep}} \label{eqn:costToGoFunction}
\end{equation}
subject to (\ref{eqn:stateTransitionFunction}), (\ref{eqn:instanteousReward}) and (\ref{eqn:policyFunction}); where $\finalTimestep$ is the time step at which the first terminal state is reached (which in general will be dependent on $\vecx_0$ and $\Weightsz$), where $\df\in [0,1]$ is a constant {\it discount factor} that specifies the relative importance of long-term costs over short term ones, and where $\expectation{\cdot}$ denotes expectation.

The function $\RpiFull{0}$ is called the {\it cost-to-go} function from state $\vecx_0$, or the {\it value function}.   

In this paper we show that when a large final impulse of cost $\rewardFinalFull{}$ is given at a terminal state $\vecx \in \TerminalStateSet$, then failure to do clipping in the final timestep of the trajectory can very significantly distort the direction of the learning gradient used by certain ADP algorithms, and thus prevent successful solution of the ADP problem.  We also show that this problem is not lessened by sampling the time steps of the underlying continuous-time process at a higher rate.  This problem affects the commonly used ADP algorithms Dual Heuristic Dynamic Programming (DHP) \cite{hic92ch13,prokhorovWunschACD}, and Backpropagation through time (BPTT) \cite{backproptime90}, both of which are described in Section \ref{sec:adpAlgorithms}, plus algorithms based on DHP such as Value-Gradient Learning \cite{fairbankAlonso2012IJCNN_vgl,fairbank2012approximating,fairbank08}.  These algorithms are all very closely related to each other \cite{
prokhorovBPPTDAC,fairbankAlonso11lorvgrtpgl}, and for purposes of explaining clipping as clearly as possible, we will use BPTT as the example.  

\begin{figure}[h]
\centering
\subfloat[][Spurious zigzag gradients can occur when clipping is not used.]{
   \scalebox{0.7}{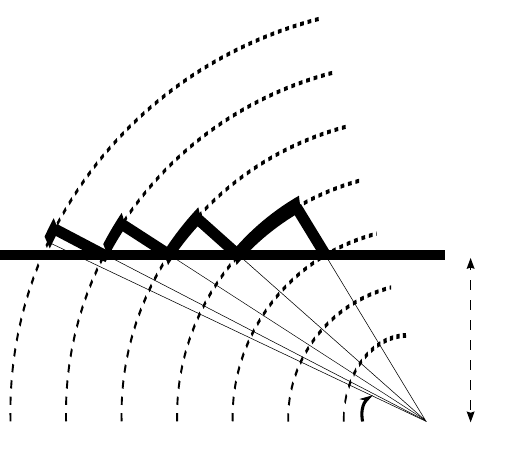}
\label{fig:subfig1}}
\qquad
\subfloat[Subfigure 2 list of figures text][The graph of $R$ versus $\theta$ yields no useful local gradient information.  Hence minimising $R$ with respect to $\theta$ using only $dR/d\theta$ would be impossible.]{
\scalebox{0.67}{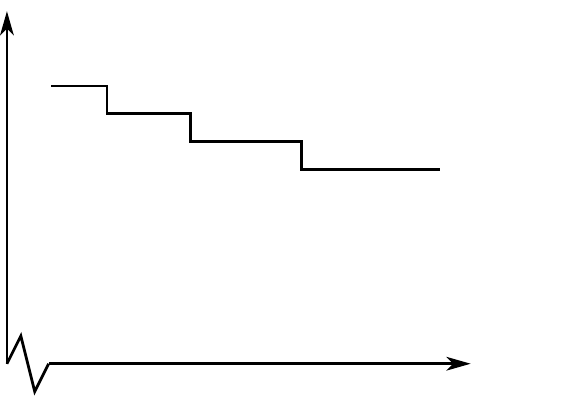}
\label{fig:subfig2}}
\qquad
\caption{An example of the problems that can occur when clipping is not used. }  
\label{fig:globfig}
\end{figure}

BPTT works by calculating the quantity $\dJdz$ directly and very efficiently for each trajectory sampled, enabling gradient descent to be performed on $J$ with respect to $\Weightsz$.  However without clipping being done correctly, the gradient that BPTT calculates can by distorted enough to prevent learning. Fig. \ref{fig:globfig} illustrates the problems that arise without clipping.

In Fig. \ref{fig:subfig1} the agent starts at $O$ and travels in a straight line at a constant speed, along a fixed chosen initial angle, $\theta$.  The straight line AB is a terminal boundary (i.e. a continuous line of states in $\TerminalStateSet$).  The dotted arcs represent the integer time steps that the agent passes through.  If clipping is not used then the agent will stop on the first integer time step (i.e. on the first dotted arc) after passing the terminal boundary.  This means the agent will finally stop at a point somewhere on the bold zigzag path from A to B.  In Fig. \ref{fig:subfig2} we see how the distance the agent travelled before stopping ($R$) varies with $\theta$.  If the cost-to-go function $J$ was defined to be the total distance travelled before termination (i.e. if $J \isDefinedToBe R$), and the parameter vector of $J$ was defined to be $\theta$, then the ADP objective would be to minimise $R$ with respect to $\theta$.  But Fig. \ref{fig:subfig2} shows that there is no useful 
gradient information for learning, since $\fracpartial{J}{\theta}=\fracpartial{R}{\theta}=0$, whenever it exists, and hence gradient descent on $J$ with respect to $\theta$ would fail without clipping.

\begin{figure}[h]
\centering
\scalebox{0.8}{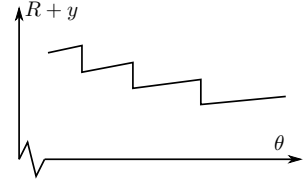}
\caption{A pathological example: Local gradient is opposite to global gradient.}
\label{fig:subfigPathalogical}
\end{figure}

Situations can get even worse than this: In Fig. \ref{fig:subfigPathalogical} we show a pathological example where the gradient of the graph is always in the opposite direction of the global minimum of $R$.  This could occur for example if we were trying to minimise the function $J\isDefinedToBe R+y$ with respect to $\theta$, for the situation in Fig. \ref{fig:subfig1}, where $y$ is the final $y$-coordinate of the agent, and $R$ is the distance travelled before stopping.

In general, increasing the sampling rate of the discretization of time will not solve the problem, since that would simply make the dotted arcs in Fig. \ref{fig:subfig1} squeeze closer together, and will make the teeth of the saw-tooth blade shape in Fig. \ref{fig:subfigPathalogical} finer.  The gradients in Figs. \ref{fig:subfig2} and \ref{fig:subfigPathalogical} would still not be helpful for learning.

We show how to solve the problem by incorporating clipping into the model and cost functions, $\modelFull{}$ and $\rewardFull{}$, when terminal states are reached.  BPTT and DHP make intensive use of the derivatives of these two functions, and hence we must carefully differentiate through the clipped versions of these functions.  This is the important step that we derive in this paper, and this step corrects the gradient $\dJdz$ to make it suitable for learning, and solves the problems explained by Fig. \ref{fig:globfig} and Fig. \ref{fig:subfigPathalogical}.

As well as terminal boundaries in state space that deliver impulses of cost, similar corrections would need making in environments where the model and cost functions change their behaviour discontinuously as the agent traverses a given continuous boundary in state space.  These boundaries would act as refraction layers do to photons. As the agent crosses them, the learning gradient $\dJdz$ would get twisted.  The solution to this problem is similar to the one we propose for terminal boundaries, but we do not consider these non-terminal refraction layers any more in this paper.

The necessity for clipping affects any algorithm that calculates the derivatives of the model function, i.e.  $\fracpartial{\model}{\vecx}$ directly, and when terminal states that deliver impulses of cost are present.  For example the RL method of \cite{munos06}, which implements a continuous-time numerical differentiation to evaluate $\dJdz$, will also be affected by this clipping problem.  Likewise, the ADP methods of BPTT, DHP, GDHP \cite{fairbank12simpleFast} and Value-Gradient Learning are also affected by the requirement for clipping.    

Clipping is  not necessary for any problem where the termination condition is simply when a fixed integer number of  time steps is reached, as we discuss further in Section \ref{sec:finiteHorizonProblems}.  Also our experiments in this paper show that the ADP algorithm called Heuristic Dual Programming (HDP, \cite{hic92ch13,adpAnIntroduction,prokhorovWunschACD}) does not need clipping,  since this algorithm does not make significant use of the derivatives of the model function.  The {\it policy-gradient learning} methods of \cite{williams,Peters06policygradient} do not require clipping either, since they do not use the derivatives of the model function.

In the rest of this paper, in Section \ref{sec:adpAlgorithms} we describe the affected ADP algorithms for control problems.  In Section \ref{sec:solution} we describe how to do the clipping and differentiate through the modified model functions, as is required for effective gradient descent. In Section \ref{sec:experimentResults} we give experimental details of neural-network control problems both with and without clipping.  One of these problems is the classic cart-pole benchmark problem which we formulate in a way that would be impossible for DHP to solve without clipping, and we show that the clipping methods enable us to solve this problem efficiently.  In Section \ref{sec:conclusions}, we give conclusions.

\section{The ADP/RL Learning Algorithms} \label{sec:adpAlgorithms}

We describe three main ADP/RL algorithms first in their forms without clipping. 

\subsection{Backpropagation Through Time For Control} \label{sec:BPTT}

Backpropagation through time (BPTT) can be applied to control problems, as described by \cite{backproptime90}. In this section we derive and describe the algorithm.  This is an algorithm that requires clipping in the environments we consider in this paper.

BPTT is an efficient algorithm to calculate $\fracpartial{\Rpi}{\Weightsz}$ for a given trajectory.  The combination of the BPTT gradient calculation with a gradient descent weight update can be used to solve control problems, i.e. by the weight update $\Delta \Weightsz=-\learningRateActor \fracpartial{\Rpi}{\Weightsz}$ for some small positive learning rate $\learningRateActor$. 

Throughout this paper we make a notational convention that all vectors are columns, and differentiation of a scalar by a vector gives a column vector (e.g. $\fracpartial{\Rpi}{\vecx}$ is a column). We define differentiation of a vector function by a vector argument as the transpose of the usual Jacobian notation.  For example, $\fracpartial{\PolicyFunctionFull{}}{\vecx}$ is a matrix with element $(i,j)$ equal to $\fracpartial{\PolicyFunction^j}{\vecx^i}$. Similarly, $\fracpartial{\model}{\vecx}$ is the matrix with element $\left(\fracpartial{\model}{\vecx}\right)^{ij}=\fracpartial{\model^j}{\vecx^i}$.  

Parentheses subscripted with a ``$t$'' are what we call {\it trajectory-shorthand notation}, which we define to indicate that a quantity is evaluated at time step $t$ of a trajectory.  For example $\fpevalat{\reward}{\veca}{t}$ is shorthand for the function $\fracpartial{\rewardFull{}}{\veca}$ evaluated at $(\vecx_t,\veca_t)$.  Similarly, $\fpevalat{\Rpi}{\vecx}{t+1} \isDefinedToBe \left. \fracpartial{\RpiFull{}}{\vecx} \right|_{(\vecx_{t+1},\Weightsz)}$, and $\fpevalat{\PolicyFunction}{\Weightsz}{t}\isDefinedToBe\left. \fracpartial{\PolicyFunctionFull{}}{\Weightsz}\right|_{(\vecx_t,\Weightsz)}$

For any given trajectory starting at state $\vecx_0$, the function $\RpiFull{0}$ given by (\ref{eqn:costToGoFunction}) can be written recursively using equations (\ref{eqn:stateTransitionFunction})-(\ref{eqn:policyFunction}), as:
\begin{align}\RpiFull{}\isDefinedToBe \reward(\vecx,\PolicyFunction(\vecx,\Weightsz))+\df \Rpi(f(\vecx,\PolicyFunction(\vecx,\Weightsz)),\Weightsz) \label{eqn:totalRewardW} \end{align}
with $\RpiFull{\finalTimestep}\isDefinedToBe \rewardFinalFull{\finalTimestep}$ at the trajectory's terminal state, $\vecx_\finalTimestep \in \TerminalStateSet$.

Differentiating (\ref{eqn:totalRewardW}) with the chain rule gives:
\allowdisplaybreaks{
\begin{small}
\begin{align}
\hspace{-1cm}
&\fpevalat{\Rpi}{\Weightsz}{t} 
& \nonumber \\ 
=& \left({ \fracpartial{}{\Weightsz} (\reward(\vecx,\PolicyFunction(\vecx,\Weightsz))+\df \Rpi(f(\vecx,\PolicyFunction(\vecx,\Weightsz)),\Weightsz))}\right)_t &\hspace{-2cm}\text{by (\ref{eqn:totalRewardW})}\nonumber \\
=& \fpevalat{\PolicyFunction}{\Weightsz}{t} \left( \fpevalat{\reward}{\veca}{t} +\df \fpevalat{f}{\veca}{t}\fpevalat{\Rpi}{\vecx}{t+1} \right) 
+\df \fpevalat{\Rpi}{\Weightsz}{t+1} &  \nonumber
\end{align}
\end{small}
where we used the chain rule,  equations (\ref{eqn:stateTransitionFunction})-(\ref{eqn:policyFunction}) and trajectory-shorthand notation.  In this equation there are implied matrix-vector products that make use of the matrix notation defined above.  

Expanding this recursion gives:
\allowdisplaybreaks{
\begin{align}
\hspace{-0in} \fpevalat{\Rpi}{\Weightsz}{0}=& \sum _{t \ge 0} \dfa{^{t}} \fpevalat{\PolicyFunction}{\Weightsz}{t} \left( \fpevalat{\reward}{\veca}{t} +\df \fpevalat{f}{\veca}{t}\fpevalat{\Rpi}{\vecx}{t+1} \right) & \label{eqn:bpttWeightUpdate}
\end{align}
}

This equation refers to the quantity $\dRdx$ which can be found recursively by differentiating (\ref{eqn:totalRewardW}) and using the chain rule, giving
\begin{align}\dRdxEvalAt{t}=&
\fpevalat{\reward}{\vecx}{t}+\gamma \fpevalat{f}{\vecx}{t}\dRdxEvalAt{t+1}
\nonumber \\ 
+&\fpevalat{\PolicyFunction}{\vecx}{t}\left(\fpevalat{\reward}{\veca}{t}+\df \fpevalat{f}{\veca}{t} \dRdxEvalAt{t+1}\right) & \label{eqn:dRdxRecursion}\end{align}
with 
\begin{equation}
\dRdxEvalAt{\finalTimestep}=\fpevalat{\rewardFinal}{\vecx}{\finalTimestep} \label{eqn:dRdx_F}
\end{equation} at the terminal state, $\vecx_\finalTimestep \in \TerminalStateSet$.

Equation (\ref{eqn:dRdxRecursion}) can be understood to be backpropagating the quantity $\dRdxEvalAt{t+1}$ through the
action network, model and cost functions to obtain  $\dRdxEvalAt{t}$, and giving the algorithm its name.  Pseudocode for the whole BPTT algorithm is given in Alg. \ref{alg:BPTT}, where lines \ref{line:algBPTT:pF}, \ref{line:algBPTT:updateDeltaZ} and \ref{line:algBPTT:updateP} of the algorithm come from equations (\ref{eqn:dRdx_F}), (\ref{eqn:bpttWeightUpdate}) and (\ref{eqn:dRdxRecursion}) respectively.  In the algorithm, the vector $\vecpInBPTTalg$ holds the backpropagated value for $\dRdx$. $\dQdx$ and $\dQdu$ are the derivatives of the Q-function with respect to $\vecx$ and $\veca$ respectively, where the Q-function is defined by
$$Q(\vecx, \veca, \Weightsz)=\rewardFull{}+\df \Rpi(\modelFull{}, \Weightsz)$$
The Q-function is a model based version of the Q-function defined in Q-learning \cite{watkins89}.  It is similar to the cost-to-go function's recursive definition (\ref{eqn:totalRewardW}), but it differs in that it allows the first action chosen to be independent of the action network.  This will be useful in deriving the clipping equations in Section \ref{sec:solution}, but for now $\dQdx$ and $\dQdu$ can just be treated as internal variables in Alg. \ref{alg:BPTT}.  The BPTT algorithm runs in time $O(\dim(\Weightsz))$ per trajectory step.

\begin{algorithm}[htp]
\begin{algorithmic}[1]
\caption {Backpropagation Through Time for Control.}
\label{alg:BPTT}
\REQUIRE {Trajectory calculated by (\ref{eqn:stateTransitionFunction}) and (\ref{eqn:policyFunction}).}
\STATE $\dJdw\leftarrow \vec{0}$
\STATE $\vecpInBPTTalg \leftarrow \fpevalat{\rewardFinal}{\vecx}{\finalTimestep}$ \label{line:algBPTT:pF}
\FOR {$t=\finalTimestep-1$ to $0$ step $-1$}
\STATE $\dQdx \leftarrow \fpevalat{\reward}{\vecx}{t}+ \gamma \fpevalat{\model}{\vecx}{t}\vecpInBPTTalg$
\STATE $\dQdu \leftarrow \fpevalat{\reward}{\veca}{t} + \gamma \fpevalat{\model}{\veca}{t}\vecpInBPTTalg$
\STATE $\dJdw \leftarrow  \dJdw+\dfa{^t} \fpevalat{\PolicyFunction}{\Weightsz}{t} 
\dQdu
$ \label{line:algBPTT:updateDeltaZ}
\STATE $\vecpInBPTTalg \leftarrow \dQdx+\fpevalat{\PolicyFunction}{\vecx}{t}\dQdu
$ \label{line:algBPTT:updateP}
\ENDFOR
\STATE $\Weightsz \leftarrow \Weightsz-\learningRateActor \dJdw$
\end{algorithmic} 
\end{algorithm}

\subsection{Dual Heuristic Dynamic Programming (DHP) and Heuristic Dynamic Programming (HDP)}

Dual Heuristic Dynamic Programming (DHP) and Heuristic Dynamic Programming (HDP) are ADP algorithms which use a critic function,  and can require clipping in the evironments we consider in this paper.  Both of these algorithms were originally by Werbos \cite{hic92ch13} and are described more recently by \cite{prokhorovWunschACD,HLADPch3Ferarri,adpAnIntroduction}, and we define them briefly here.  

The use of critic functions allows these two algorithms to apply their learning rule on-line, unlike the previously described BPTT which needed to wait until a trajectory was completed before it could apply the learning weight update.  DHP makes use of a {\it vector-critic} function $\GapproxFull{}$ which produces a vector output of dimension $\Reals^{\dim(\vecx)}$.  This could be the output of a neural network with weight vector $\Weights$ and $\dim(\vecx)$ inputs and outputs.  The DHP weight update attempts to make the function $\GapproxFull{}$ learn to output the gradient $\fracpartial{\Rpi}{\vecx}$.   HDP uses a {\it scalar-critic} function $\VapproxFull{}$ which produces a scalar output.  This could be the output of a neural network with weight vector $\Weights$ and $\dim(\vecx)$ inputs, and just one output node.  The HDP weight update attempts to make the function $\VapproxFull{}$ learn to output the function $\RpiFull{}$ for all $\vecx \in \StateSpace$.  HDP is equivalent to the algorithm ``TD(0)'' 
from the RL literature \cite{sutton88learning}.

Pseudocode for DHP is given in Alg. \ref{alg:DHP}.  Line \ref{line:DHP:criticWeightUpdate} of the algorithm trains the critic with a learning rate $\learningRateCritic>0$, and line \ref{line:DHP:actorWeightUpdate} implements a commonly used actor weight update described by \cite{prokhorovWunschACD} (using a learning rate $\learningRateActor>0$).  The algorithm uses the same matrix notation for Jacobians and trajectory-shorthand notation as described in Section \ref{sec:BPTT}, so that for example $\fpevalat{\Gapprox}{\Weights}{t}$ is the function $\fracpartial{\Gapprox}{\Weights}$ evaluated at $(\vecx_t, \Weights)$.

Pseudocode for HDP is given in Alg. \ref{alg:HDP}.  Lines \ref{line:HDP:criticWeightUpdate} and \ref{line:HDP:actorWeightUpdate} give the critic and action-network weight updates, respectively.  Again the action-network weight update is the one described by \cite{prokhorovWunschACD}, but model-free alternatives which don't require knowledge of the derivatives of $\model$ are also possible (e.g. \cite[ch.6.6]{suttonbarto-1998}, or \cite[sec 4.2]{doya00reinforcement}).

Backpropagation (\cite{werbos1974,RumelhartHintonWilliams1986}) can be used to efficiently calculate $\fracpartial{\Vapprox}{\Weights}$, $\fracpartial{\Vapprox}{\vecx}$ and the products involving $\fracpartial{\PolicyFunction}{\Weightsz}$ and $\fracpartial{\Gapprox}{\Weights}$.  Using this method, both  DHP and HDP can be implemented in a running time of $O(n)$ operations per time step of the trajectory, where $n=\max(\dim(\Weights),\dim(\Weightsz))$.

The pseudocode gives explicit details of how the function $\rewardFinalFull{}$ is to be used instead of the critic at the final time step of a trajectory.  This is an important detail that is necessary to implement clipping and finite-horizon problems correctly.

\begin{algorithm}[h]
\begin{algorithmic}[1]
\caption {DHP with a critic network $\GapproxFull{}$ and action network $\PolicyFunctionFull{}$.}
\label{alg:DHP}
\STATE $t\leftarrow 0$
\WHILE{$\vecx_{t} \notin \TerminalStateSet$} 
\STATE $\veca_t \leftarrow \PolicyFunctionFull{t}$
\STATE $\vecx_{t+1} \leftarrow \modelFull{t}$ \label{line:DHP:nextStateEvaluation}
\STATE $\vecpInBPTTalg \leftarrow \begin{cases}
                            \fpevalat{\rewardFinal}{\vecx}{t+1} & \text{if $\vecx_{t+1} \in \TerminalStateSet$} \\
			    \GapproxFull{t+1}	& \text{if $\vecx_{t+1} \notin \TerminalStateSet$}
                           \end{cases}$ \label{line:dhp:vecpAssignment}
\STATE $\dQdx \leftarrow \fpevalat{\reward}{\vecx}{t}+ \gamma \fpevalat{\model}{\vecx}{t}\vecpInBPTTalg$
\STATE $\dQdu \leftarrow \fpevalat{\reward}{\veca}{t} + \gamma \fpevalat{\model}{\veca}{t}\vecpInBPTTalg$
\STATE $\vecDelta \leftarrow \dQdx
 +\fpevalat{\PolicyFunction}{\vecx}{t}\dQdu-\GapproxFull{t}$
\STATE $\Weights \leftarrow  \Weights+ \learningRateCritic \fpevalat{\Gapprox}{\Weights}{t} \vecDelta$ \label{line:DHP:criticWeightUpdate} \COMMENT {Critic network update}
\STATE $\Weightsz \leftarrow  \Weightsz-\learningRateActor \fpevalat{\PolicyFunction}{\Weightsz}{t}\dQdu$ \label{line:DHP:actorWeightUpdate}\COMMENT {Action network update}
\STATE $t \leftarrow t+1$
\ENDWHILE
\end{algorithmic} 
\end{algorithm}

\begin{algorithm}[h]
\begin{algorithmic}[1]
\caption {HDP with a critic network $\VapproxFull{}$ and action network $\PolicyFunctionFull{}$}
\label{alg:HDP}
\STATE $t\leftarrow 0$
\WHILE{$\vecx_{t} \notin \TerminalStateSet$} 
\STATE $\veca_t \leftarrow \PolicyFunctionFull{t}$
\STATE $\vecx_{t+1} \leftarrow \modelFull{t}$
\STATE $\vecpInBPTTalg \leftarrow \begin{cases}
                            \fpevalat{\rewardFinal}{\vecx}{t+1} & \text{if $\vecx_{t+1} \in \TerminalStateSet$} \\
			    \fpevalat{\Vapprox}{\vecx}{t+1}	& \text{if $\vecx_{t+1} \notin \TerminalStateSet$}
                           \end{cases}$ \label{line:hdp:vecpAssignment}
\STATE $\hdpTarget \leftarrow \begin{cases}
                            \rewardFinalFull{t+1} & \text{if $\vecx_{t+1} \in \TerminalStateSet$} \\
			    \VapproxFull{t+1}	& \text{if $\vecx_{t+1} \notin \TerminalStateSet$}
                           \end{cases}$ \label{line:hdp:hdptargetAssignment}
\STATE $\dQdu \leftarrow \fpevalat{\reward}{\veca}{t} + \gamma \fpevalat{\model}{\veca}{t}\vecpInBPTTalg$
\STATE $\Weights \leftarrow  \Weights+ \learningRateCritic \fpevalat{\Vapprox}{\Weights}{t} \left(\rewardFull{t}+ \gamma \hdpTarget-\VapproxFull{t}\right)$ \label{line:HDP:criticWeightUpdate} \COMMENT {Critic network update}
\STATE $  \Weightsz \leftarrow  \Weightsz-\learningRateActor \fpevalat{\PolicyFunction}{\Weightsz}{t}\dQdu$ \label{line:HDP:actorWeightUpdate}\COMMENT {Action network update}
\STATE $t \leftarrow t+1$ 
\ENDWHILE
\end{algorithmic} 
\end{algorithm}

\section{Using and Differentiating Clipping in Learning} \label{sec:solution}
In this section we derive the formulae for the clipped model and cost functions, and their derivatives.  We will denote the clipped versions of the original functions with a superscripted {\it C}, so that  $\modelClipped$, $\rewardClipped$ and $\RpiClipped$ will be the function names we use for the clipped versions of the model, cost and cost-to-go functions, respectively.  The functions $\modelClipped$ and $\rewardClipped$ are only defined for any state $\vecx_t$ that occurs immediately {\it before} a terminal state is reached, i.e. for which $\vecx_t \notin \TerminalStateSet$ and for which $\modelFull{t}\in \TerminalStateSet$.  

These three clipped functions, $\modelClipped$, $\rewardClipped$ and $\RpiClipped$, are key concepts in this paper, because defining them clearly allows us to differentiate them carefully, and hence calculate the learning gradients correctly.  This is what allows us to solve the clipping problem.  Hence this section is the main contribution of this paper, in terms of implementation details for solving the clipping problem.

\subsection{Calculation of the Clipped Model and Cost Functions}

Suppose the agent is transitioning between states $\vecx_t$ and $\modelFull{t}$, and the state $\modelFull{t}$ would be beyond the terminal boundary unless clipping was applied.  To calculate the clipping correctly, we imagine this state transition as occurring along the straight line segment from $\vecx_t$ to $\modelFull{t}$, i.e. the straight line given parametrically by position vector 
\begin{equation}
\vecr=\vecx_t+\lineParameter \vecv, \label{eqn:straightLine}
\end{equation}where 
\begin{equation}
\vecv=\modelFull{t}-\vecx_t, \label{eqn:vecv}
\end{equation} and  $\lineParameter \in [0,1]$ is a real parameter.  This is illustrated in Fig. \ref{fig:clippingLinePlane}.

This straight line must intersect a boundary of terminal states.  At the point of intersection, the tangent plane of the terminal boundary is given by $(\vecr-\planeConstantP)\cdot\vecn=0$ (i.e. where $\vecr$ is an arbitrary position vector that lies on a plane which has normal $\vecn$ and passes through a point with position vector $\planeConstantP$, and where ``$\cdot$'' denotes the inner product between two vectors), as illustrated in Fig. \ref{fig:clippingLinePlane}.  The constants $\planeConstantP$ and $\vecn$ should be available from either the physical environment or from the collision-detection routine of the simulated environment.

 \begin{figure} 
   \centering
   \def\svgwidth{\columnwidth} 
   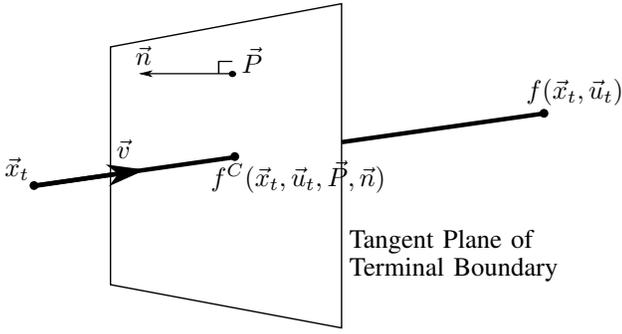
  \caption{The final state transition of a trajectory crossing the tangent plane of a terminal boundary.  The unclipped line goes from $\vecx_t$ to $\modelFull{t}$.  The line intersects the plane at a point given by the new {\it clipped} model function $\modelClippedFull{t}$.}
\label{fig:clippingLinePlane}
 \end{figure}

At the intersection of the line and the plane, we have 
\begin{align*}
& (\vecx_t+\lineParameter \vecv-\planeConstantP)\cdot \vecn=0 &\\
\Rightarrow &\lineParameter=\frac{(\planeConstantP-\vecx_t) \cdot \vecn}{\vecv\cdot\vecn}. &
\end{align*}

This value of $\lineParameter$ is a real number between 0 and 1 which indicates the fraction along the transition line from $\vecx_t$ to $\modelFull{t}$ at which the terminal boundary was encountered.  We will refer to the value $\lineParameter$ as the ``clipping fraction'', and since it depends on $\vecx_t$, $\veca_t$, $\planeConstantP$ and $\vecn$, it is defined by the function:
\begin{equation}
 \lineParameter \isDefinedToBe \clippingFractionFull{t} \isDefinedToBe \frac{(\planeConstantP-\vecx_t)\cdot\vecn}{(\modelFull{t}-\vecx_t)\cdot\vecn}. \label{eqn:clippingFractionEquation}
\end{equation}

Hence the clipped value of the final state is $\vecx_{t+1}=\vecx_t+\clippingFractionFull{t}(\modelFull{t}-\vecx_t)$, which is found by combining equations (\ref{eqn:straightLine}), (\ref{eqn:vecv}) and (\ref{eqn:clippingFractionEquation}).  This gives the function for the clipped model function as
\begin{equation}
 \modelClippedFull{} \isDefinedToBe\vecx+\clippingFractionFull{}(\modelFull{}-\vecx). \label{eqn:clippedModelFunction}
\end{equation}

Assuming that ``cost'' is delivered at a uniform rate during the final state transition, the total clipped cost would be proportional to the clipping fraction, giving:
\begin{equation}
 \rewardClippedFull{} \isDefinedToBe\clippingFractionFull{}\rewardFull{}. \label{eqn:clippedRewardFunction}
\end{equation}

Since the final clipped timestep has duration $\lineParameter \in [0,1]$, the terminal cost $\rewardFinalFull{\finalTimestep}$ should only receive a discount of $\dfa{^\lineParameter}$ instead of the full discount $\df$.  Hence, at the penultimate time step, $\vecx_{\finalTimestep-1}$, the total cost-to-go is 
\begin{align}
\RpiClippedFull{\finalTimestep-1}\isDefinedToBe 
 \rewardClippedFull{\finalTimestep-1}+\dfa{^\lineParameter}\rewardFinalFull{\finalTimestep}.
 \label{eqn:clippedCostToGoFunctionPenultimateTimestep}
\end{align}

This possibly seems like a pedantic detail, but it is this detail which allows us to solve a version of the cart-pole benchmark problem, which would otherwise be impossible for DHP, in Section \ref{sec:cartPoleExpt}.

Alg. \ref{alg:unrollClippedTrajectory} illustrates how equations (\ref{eqn:stateTransitionFunction})-(\ref{eqn:policyFunction}) and (\ref{eqn:clippingFractionEquation})-(\ref{eqn:clippedCostToGoFunctionPenultimateTimestep}) would be used to evaluate a trajectory with clipping.

\begin{algorithm}[h]
\begin{algorithmic}[1]
\caption {Unrolling a trajectory with clipping.}
\label{alg:unrollClippedTrajectory}
\STATE $t\leftarrow 0$, $\RpiClipped\leftarrow0$
\WHILE {$\vecx_{t} \notin \TerminalStateSet$}
\STATE $\veca_t \leftarrow \PolicyFunctionFull{t}$
\STATE $\vecx_{t+1} \leftarrow \modelFull{t}$  \label{line:unrollClippedTrajectory:nextStateEvaluationStart}
\IF{$\vecx_{t+1} \in \TerminalStateSet$} \label{line:trajectoryClipped:collisionDetection}
\STATE {Identify $\planeConstantP$ and $\vecn$ by inspection of the intersection with the terminal boundary, $\TerminalStateSet$.} \label{line:trajectoryClipped:determinationOfPandN}
\STATE $\lineParameter \leftarrow \clippingFractionFull{t}$
\STATE $\finalTimestep\leftarrow t+1$
\STATE $\vecx_{\finalTimestep} \leftarrow \vecx_t+\lineParameter \left(\vecx_{\finalTimestep}-\vecx_t\right)$
\STATE $\RpiClipped \leftarrow \RpiClipped+\left(\df^t\right)\left(\lineParameter \rewardFull{t}+\df^{\lineParameter}\rewardFinalFull{\finalTimestep}\right)$
\ELSE
\STATE $\RpiClipped \leftarrow \RpiClipped+\left(\df^t\right)\rewardFull{t}$
\ENDIF  \label{line:unrollClippedTrajectory:nextStateEvaluationEnd}
\STATE $t \leftarrow t+1$
\ENDWHILE
\STATE $\finalTimestep \leftarrow t$
\end{algorithmic} 
\end{algorithm}

Note that $\planeConstantP$ and $\vecn$ are required by equations (\ref{eqn:clippingFractionEquation})-(\ref{eqn:clippedRewardFunction}).  These would be found during the collision-detection routine (i.e. line \ref{line:trajectoryClipped:determinationOfPandN} of Alg. \ref{alg:unrollClippedTrajectory}), from knowledge of the terminal-boundary orientation, together with knowledge of $\vecx_{\finalTimestep-1}$ and $\modelFull{\finalTimestep-1}$.  Knowledge of the orientation of the terminal boundary could come from a model of the physical environment's boundary; or if this model was not available, then a physical inspection of the actual boundary would need to take place.  Examples of how these two vectors were found in our experiments are given in Sec. \ref{sec:oneDLLExperiment} and \ref{sec:cartPoleExpt}.  

\subsection{Calculation of the Derivatives of the Clipped Model and Cost Functions} 

The ADP algorithms described in Section \ref{sec:adpAlgorithms} require the derivatives of the model function, and hence they will require the derivatives of the clipped model function $\modelClippedFull{}$ too.  Fig. \ref{fig:endPointCorrection} shows how different the derivative of $\modelClipped$ can be from the derivative of $\model$, and hence how important it is to get this correct in ADP/RL.  This figure clarifies why algorithms that are dependent on $\fracpartial{\modelClipped}{\vecx}$ are critically affected by the need for clipping, and also that just reducing the duration of each time step tracking or simulating the motion will not solve the problem at all.

 \begin{figure} 
   \centering
   \def\svgwidth{\columnwidth} 
   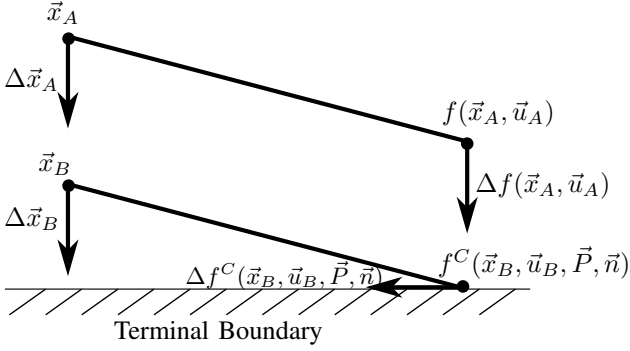
  \caption{This diagram shows how the derivatives of the model function $\modelFull{}$ radically change as the agent approaches a terminal boundary.   The straight line segment from $\vecx_A$ to $\modelFull{A}$ represents a state transition that is not intersecting the terminal boundary.  If the start of this line segment is perturbed in the direction of the arrow $\Delta \vecx_A$ then its other end will move in the direction indicated by the arrow  $\Delta \modelFull{A}$. The line segment below, however, which starts at $\vecx_B$, does reach the terminal boundary.  If the start of this line segment is moved in the direction of $\Delta \vecx_B$, then its end will move in a perpendicular direction, as indicated by the arrow $\Delta \modelClippedFull{B}$.  This indicates that $\fpevalat{\modelClipped}{\vecx}{A}$ is very different from $\fpevalat{\model}{\vecx}{B}$, and hence this needs treating carefully in the ADP algorithms.}
\label{fig:endPointCorrection}
 \end{figure}

Differentiating the formula for $\clippingFractionFull{}$ in (\ref{eqn:clippingFractionEquation}) gives:
\begin{align}
\fracpartial{\clippingFractionFull{}}{\vecx}&=\fracpartial{}{\vecx}\left(\frac{(\planeConstantP-\vecx)\cdot\vecn}{(f(\vecx, \veca)-\vecx)\cdot\vecn}\right) &\hspace{-8cm}\text{by (\ref{eqn:clippingFractionEquation})}\nonumber \\       
&=\frac{-\vecn}{\vecv\cdot\vecn}-\frac{(\planeConstantP-\vecx)\cdot\vecn}{(\vecv\cdot\vecn)^2}\fracpartial{(f(\vecx, \veca)-\vecx )\cdot\vecn}{\vecx} \hspace{-0cm}&\nonumber \\ &&\hspace{-8cm}\text{using (\ref{eqn:vecv})}
\nonumber \\       
&=\frac{-\vecn}{\vecv\cdot\vecn}-\frac{(\planeConstantP-\vecx)\cdot\vecn}{(\vecv\cdot\vecn)^2}\left(\fracpartial{f}{\vecx}-I \right)\vecn & \label{eqn:dCF_dx}
\end{align}
where $I$ is the identity matrix, and the matrix notation is as defined in Section \ref{sec:BPTT}.  Similarly,
\begin{align}
\fracpartial{\clippingFractionFull{}}{\veca}&=\fracpartial{}{\veca}\left(\frac{(\planeConstantP-\vecx)\cdot\vecn}{(f(\vecx, \veca)-\vecx )\cdot\vecn}\right) &\text{by (\ref{eqn:clippingFractionEquation})}\nonumber \\       
=&-\frac{(\planeConstantP-\vecx)\cdot\vecn}{(\vecv\cdot\vecn)^2}\fracpartial{(f(\vecx, \veca)-\vecx)\cdot\vecn}{\veca}&\text{using (\ref{eqn:vecv})} \nonumber \\       
=&-\frac{(\planeConstantP-\vecx_)\cdot\vecn}{(\vecv\cdot\vecn)^2}\left(\fracpartial{f}{\veca}\right)\vecn  &  \label{eqn:dCF_du}
\end{align}

Using these derivatives of $\clippingFractionFull{}$, we can now differentiate the clipped model and cost functions, giving:
\begin{align}
 \fracpartial{\modelClippedFull{}}{\vecx}&=I+\fracpartial{\clippingFraction}{\vecx}\vecv^T+\lineParameter \left(\fracpartial{f}{\vecx}-I\right) & \nonumber \\
&&\hspace{-3cm}\text{by (\ref{eqn:vecv})-(\ref{eqn:clippedModelFunction})}\label{eqn:dfClipped_dx} \\
 \fracpartial{\modelClippedFull{}}{\veca}&=\fracpartial{\clippingFraction}{\veca}\vecv^T+\lineParameter \fracpartial{f}{\veca} & \hspace{-2cm} \text{by (\ref{eqn:vecv})-(\ref{eqn:clippedModelFunction})} \label{eqn:dfClipped_du} \\
 \fracpartial{\rewardClippedFull{}}{\vecx}&=\fracpartial{\clippingFraction}{\vecx}\rewardFull{}+\lineParameter \fracpartial{\reward}{\vecx} & \hspace{-0.5cm}\text{by (\ref{eqn:clippedRewardFunction})} \label{eqn:dUClipped_dx} \\
 \fracpartial{\rewardClippedFull{}}{\veca}&=\fracpartial{\clippingFraction}{\veca}\rewardFull{}+\lineParameter \fracpartial{\reward}{\veca} & \hspace{-0.5cm}\text{by (\ref{eqn:clippedRewardFunction})}\label{eqn:dUClipped_du} 
\end{align}

The cost-to-go function for the penultimate time step, equation (\ref{eqn:clippedCostToGoFunctionPenultimateTimestep}), can be rewritten as a Q-function of both $\vecx$ and $\veca$, to give
\begin{align}Q(\vecx_{\finalTimestep-1}, \veca_{\finalTimestep-1}) \isDefinedToBe &\rewardClippedFull{\finalTimestep-1}\nonumber \\&+\df^{\lineParameter}\rewardFinal(\modelClippedFull{\finalTimestep-1}).\label{eqn:discountedFinalQFunctionClipped}\end{align}  
Differentiating this with respect to $\veca_{\finalTimestep-1}$ or $\vecx_{\finalTimestep-1}$ gives:
\begin{align}
 \fpevalat{Q}{\bullet}{\finalTimestep-1} =& \fpevalat{\rewardClipped}{\bullet}{\finalTimestep-1}+\dfa{^{\lineParameter}}\left(\fpevalat{\modelClipped}{\bullet}{\finalTimestep-1}\fpevalat{\rewardFinal}{\vecx}{\finalTimestep} 
\right.\nonumber\\&+\left. 
\left(\ln{\df}\right)\fpevalat{\clippingFraction}{\bullet}{\finalTimestep-1}\rewardFinalFull{\finalTimestep}\right) \label{eqn:derivative_DiscountedFinalCostClipped}
\end{align}
where $\bullet$ represents either $\veca$ or $\vecx$.

This equation, which relies upon the derivatives of $\modelClippedFull{}$ and $\rewardClippedFull{}$ (as defined in equations (\ref{eqn:dCF_dx}) to (\ref{eqn:dUClipped_du})), can be used to modify BPTT from Alg. \ref{alg:BPTT} into its corresponding ``with clipping'' version given in Alg. \ref{alg:BPTTwithClipping}. Equation (\ref{eqn:derivative_DiscountedFinalCostClipped}) appears in the algorithm directly in lines \ref{line:BPTTwithClipping:dQdx}-\ref{line:BPTTwithClipping:dQdu}.

The DHP and HDP algorithms need similar modifications to convert them to include clipping. Clipping needs applying to the final time step of the trajectory unroll, which can be implemented by replacing line \ref{line:DHP:nextStateEvaluation} of both algorithms by lines \ref{line:unrollClippedTrajectory:nextStateEvaluationStart}-\ref{line:unrollClippedTrajectory:nextStateEvaluationEnd} of Alg. \ref{alg:unrollClippedTrajectory}.  Also, in the case of DHP (Alg. \ref{alg:DHP}), the lines that calculate $\dQdx$ and $\dQdu$ need replacing by lines \ref{line:BPTTwithClipping:calcQderivsStart}-\ref{line:BPTTwithClipping:calcQderivsEnd} of Alg. \ref{alg:BPTTwithClipping}; and similarly the line that calculates $\dQdu$ in Alg. \ref{alg:HDP} (HDP) needs the same modification.

\begin{algorithm}[htp]
\begin{algorithmic}[1]
\caption {Backpropagation Through Time for Control, with Clipping.}
\label{alg:BPTTwithClipping}
\REQUIRE {Trajectory calculated by Alg. \ref{alg:unrollClippedTrajectory}}
\STATE $\dJdw\leftarrow \vec{0}$
\STATE $\vecpInBPTTalg \leftarrow \fpevalat{\rewardFinal}{\vecx}{\finalTimestep}$ 
\FOR {$t=\finalTimestep-1$ to $0$ step $-1$}
\IF{$\vecx_{t+1} \in \TerminalStateSet$} \label{line:BPTTwithClipping:calcQderivsStart}
\STATE {Calculate $\fpevalat{\clippingFraction}{\vecx}{t}$ and $\fpevalat{\clippingFraction}{\veca}{t}$ by (\ref{eqn:dCF_dx}) and (\ref{eqn:dCF_du}).}
\STATE {Calculate $\fpevalat{\modelClipped}{\vecx}{t}$ and $\fpevalat{\modelClipped}{\veca}{t}$ by (\ref{eqn:dfClipped_dx}) and (\ref{eqn:dfClipped_du}).}
\STATE {Calculate $\fpevalat{\rewardClipped}{\vecx}{t}$ and $\fpevalat{\rewardClipped}{\veca}{t}$ by (\ref{eqn:dUClipped_dx}) and (\ref{eqn:dUClipped_du}).}
\STATE $\dQdx \leftarrow 
\fpevalat{\rewardClipped}{\vecx}{t}$\\$
\ \ \ \ \ \ + \gamma^{\lineParameter} \left(\fpevalat{\modelClipped}{\vecx}{t}\vecpInBPTTalg+\left(\ln \gamma\right)\fpevalat{\clippingFraction}{\vecx}{t}\rewardFinalFull{\finalTimestep}\right)$ \label{line:BPTTwithClipping:dQdx}
\STATE $\dQdu \leftarrow \fpevalat{\rewardClipped}{\veca}{t}
$\\$
\ \ \ \ \ \ + \gamma^{\lineParameter} \left(\fpevalat{\modelClipped}{\veca}{t}\vecpInBPTTalg
+\left(\ln \gamma\right)\fpevalat{\clippingFraction}{\veca}{t}\rewardFinalFull{\finalTimestep}\right)$ \label{line:BPTTwithClipping:dQdu}
\ELSE
\STATE $\dQdx \leftarrow \fpevalat{\reward}{\vecx}{t}+ \gamma \fpevalat{\model}{\vecx}{t}\vecpInBPTTalg$
\STATE $\dQdu \leftarrow \fpevalat{\reward}{\veca}{t} + \gamma \fpevalat{\model}{\veca}{t}\vecpInBPTTalg$
\ENDIF \label{line:BPTTwithClipping:calcQderivsEnd}
\STATE $\dJdw \leftarrow  \dJdw+\dfa{^t} \fpevalat{\PolicyFunction}{\Weightsz}{t} 
\dQdu
$
\STATE $\vecpInBPTTalg \leftarrow \dQdx+\fpevalat{\PolicyFunction}{\vecx}{t}\dQdu
$ 
\ENDFOR
\STATE $\Weightsz \leftarrow \Weightsz-\learningRateActor \dJdw$
\end{algorithmic} 
\end{algorithm}

\subsection{Implementing Clipping Efficiently and Correctly}

To demonstrate how clipping would be correctly implemented with an ADP/RL algorithm, we use the BPTT algorithm for illustration.  In an implementation of BPTT with clipping, we would first evaluate a trajectory by Alg. \ref{alg:unrollClippedTrajectory}.  During this stage, we would record the full trajectory $(\vecx_0, \vecx_1, \ldots, \vecx_\finalTimestep)$ and actions $(\veca_0, \veca_1, \ldots, \veca_{\finalTimestep-1})$ and also, during the collision with the terminal boundary, we would record $\planeConstantP$ and $\vecn$ and the clipping fraction, $\lineParameter$.  We then have enough information to be able to run the BPTT algorithm with clipping (Alg. \ref{alg:BPTTwithClipping}).

To ensure the correctness of our implementations in each experiment and environment which we tackled, we first verified all of the derivatives of $\clippingFractionFull{}$, $\modelClippedFull{}$ and $\rewardClippedFull{}$ numerically, with respect to both $\vecx$ and $\veca$, at least a few times.  When all of these derivatives were all satisfactorily programmed and checked, we then checked by numerical differentiation that the overall BPTT implementation was calculating the derivative $\fracpartial{\RpiClipped}{\Weightsz}$ correctly.  

For an example of the numerical differentiations used, the final check of BPTT was done by a central-differences numerical derivative for each component $i$ of the weight vector $\Weightsz$, to verify that
$$\fracpartial{{\RpiClipped}}{\Weightsz^i} = \frac{{\RpiClipped}(\vecx_0, \Weightsz+ \epsilon \vec{e}_i)-{\RpiClipped}(\vecx_0, \Weightsz- \epsilon \vec{e}_i)}{2\epsilon}+O(\epsilon^2)$$
where $\epsilon$ is a small positive constant, and $\vec{e}_i$ is the $i$th Euclidean standard basis vector.  In this verification equation, each ${\RpiClipped}(\cdot)$ term appearing in the right-hand side would be computed by executing Alg. \ref{alg:unrollClippedTrajectory} from the trajectory start point $\vecx_0$; and the theoretical value of $\fracpartial{\RpiClipped}{\Weightsz}$ appearing in the left-hand side would be computed by Alg. \ref{alg:BPTTwithClipping}.

In HDP and DHP, the derivatives of $\clippingFractionFull{}$, $\modelClippedFull{}$ and $\rewardClippedFull{}$ would be calculated and verified as above.  However with HDP and DHP it is more difficult to check the overall critic weight updates numerically, since they are not true gradient descent on any analytic function \cite{barnard93}.    For these algorithms, it was possible to verify the key algorithmic modifications related to clipping, by just checking the derivatives of the Q-function given by (\ref{eqn:derivative_DiscountedFinalCostClipped}).  These derivatives were compared to the numerical derivatives of (\ref{eqn:discountedFinalQFunctionClipped}) with respect to $\vecx$ and $\veca$.

\subsection{Clipping with Trajectories of Fixed or Variable Finite Length} \label{sec:finiteHorizonProblems}

In situations where trajectories are fixed finite length (commonly referred to as a fixed-length finite-horizon problem),  clipping is not necessary. This is in contrast to the problems we considered in the introduction, which were variable finite-length problems, since the trajectory lengths were determined by the environment (e.g. a trajectory terminates only when the agent crashes into a wall).  In this section we will distinguish between these two situations by referring to them as ``fixed finite-length'' and ``variable finite-length'' problems, respectively.  Only in variable finite-length problems is clipping necessary.

In the fixed finite-length problem, the clipping fraction defined by (\ref{eqn:clippingFractionEquation}) is always $\lineParameter \equiv 1$, and therefore $\fracpartial{\clippingFraction}{\vecx} = \vec{0}$,  $\fracpartial{\clippingFraction}{\veca} = \vec{0}$ and $\df^\lineParameter = \df$.  Hence the clipped model and cost functions are identical to their unclipped counterparts, and therefore it is not necessary to implement any program code specifically to handle clipping.  This might be one reason why the need for clipping has not previously been noted in the research literature, since most finite-horizon problems considered have been fixed-finite length.

However the fixed finite-length problem does have one minor different complication, in that it is often necessary to include the time step into the state vector.  This is because the optimal actions and cost-to-go function will often be dependent upon the number of incomplete steps in a trajectory. 

Of course for both fixed-length and variable-length finite-horizon problems, it is important to ensure the terminal cost function $\rewardFinalFull{}$ is learnt correctly by the learning algorithm.  The pseudocode shows explicitly how to do this (e.g. for BPTT, see line \ref{line:algBPTT:pF} of Algs. \ref{alg:BPTT} and \ref{alg:BPTTwithClipping}. For DHP, see line  \ref{line:dhp:vecpAssignment} of Alg. \ref{alg:DHP}. And for HDP, see lines \ref{line:hdp:vecpAssignment} and \ref{line:hdp:hdptargetAssignment} of Alg. \ref{alg:HDP}).  

\section{Experimental Results} \label{sec:experimentResults}

This section describes two neural-network based ADP/RL experiments which require clipping to be solved well.

In all experiments the action and critic networks used were multi-layer perceptrons (MLPs, see \cite{bishop95} for details).  Each MLP had $\dim(\vecx)$ input nodes, 2 hidden layers of 6 nodes each, and one output layer, with short-cut connections connecting all pairs of layers.   The output layers were dimensioned as follows:  Each action network had $\dim(\veca)$ output nodes; each HDP critic network had 1 output node; and each DHP critic had $\dim(\vecx)$ output nodes.  All network nodes had bias weights, as is usual in MLP architectures.  The activation functions used were hyperbolic tangent functions, except for the critic network's output layer which was always a linear activation function (with linear slope as specified in the individual experiments, below).  At the start of each experimental trial, neural weights were initialised randomly in the range $[-.1,.1]$, with uniform probability distribution.

\subsection{Vertical Lander problem} \label{sec:oneDLLExperiment}

A spacecraft is dropped in a uniform gravitational field, and its objective is to make a fuel-efficient gentle landing.  The spacecraft is constrained to move in a vertical line, and a single thruster is available to make upward accelerations.  The state vector $\vecx =(h, v, u)^T$ has three components: height ($h$), velocity ($v$), and fuel remaining ($u$).  The action vector, $a$, is one-dimensional (so that $\veca \equiv a \in \Reals$) producing accelerations  $a \in [0,1]$. The Euler method with time-step $\Delta t$ is used to integrate the motion, giving model functions:
\begin{align}
f((h, v, u)^T, a)=&(h+v\Delta t, v+(a-k_g)\Delta t, (k_u)u-a\Delta t)^T \nonumber \\
\reward((h, v, u)^T, a)=&(k_f) a \Delta t  \label{eqn:LLModelFunctionsR}
\end{align}

Here, $k_g=0.2$ is a constant giving the acceleration due to gravity;  the spacecraft can produce greater acceleration than that due to gravity.  $k_f=4$ is a constant giving fuel penalty.  $k_u=1$ is a unit conversion constant.  We used $\Delta t=1$ in our main experiments here.

Trajectories terminate as soon as the spacecraft hits the ground ($h=0$) or runs out of fuel ($u=0$).  These two conditions define $\TerminalStateSet$.  This is a variable finite-length problem, and there is no need to use a discount factor, so we fixed $\df=1$.  On termination, the algorithms need to choose values for $\planeConstantP$, and $\vecn$ which describe the orientation of the terminal-boundary tangent plane.  These choices are given for this experiment in Table \ref{table:terminationBoundaryPlanesLL}.  In the case that the final un-clipped state transition crosses both terminal planes, then the one that is crossed first (i.e the one that produces a smaller clipping fraction by (\ref{eqn:clippingFractionEquation})) is to be used.

In addition to the cost function $\reward(\vecx, a)$ defined above, a final impulse of cost defined by $\rewardFinalFull{\finalTimestep}\isDefinedToBe \half m v^2+m(k_g)h$ is given as soon as the lander reaches a terminal state, where $m=2$ is the mass of the spacecraft.  The two terms in the final impulse of cost are the kinetic and potential energy, respectively.  The first cost term penalises landing too quickly. The second term is a cost term equivalent to the kinetic energy that the spacecraft would acquire by crashing to the ground under free fall (i.e. with $a=0$), so to minimise this cost the spacecraft must learn to not run out of fuel.

The input vector to the action and critic networks was $\vecx'=\left(h/100, v/10, u/50\right)^T$, and the model and cost functions were redefined to act on this rescaled input vector directly.   The action network's output $y$ was rescaled to give the action by $\PolicyFunctionFull{} \isDefinedToBe (y+1)/2$ directly.  We tested each algorithm in batch mode, operating on five trajectories simultaneously.  Those five trajectories had fixed start points, which had been randomly chosen in the region $h\in (0,100)$, $v \in (-10,10)$ and $u=30$.  

Fig. \ref{fig:LL_performance} shows learning performance of the BPTT, DHP and HDP algorithms, both with and without clipping.  Each graph shows five curves, and each curve shows the learning performance from a different random weight initialisation.  The learning rates for the three algorithms were: BPTT ($\learningRateActor=0.01$); DHP ($\learningRateActor=0.001$, $\learningRateCritic=0.00001$); and HDP ($\learningRateActor=0.00001$, $\learningRateCritic=0.00001$).  The critic-network's output layer's activation function had a linear slope of 20 in the DHP experiment and 10 in the HDP experiment.

Because HDP is an algorithm which requires stochastic exploration to optimise the ADP/RL problem effectively \cite{fairbankAlonso2012IJCNN_dhpTdComparison}, in the HDP experiment we had to modify (\ref{eqn:policyFunction}) to choose exploratory actions.  Hence for the HDP experiment we used  
$$\veca_t=\PolicyFunctionFull{t}+X_\sigma,$$
where $X_\sigma$ is a normally distributed random variable with mean zero and standard deviation $\sigma=0.1$.

These graphs show the clear stability and performance advantages of using clipping correctly for the BPTT and DHP algorithms.  The graphs also confirm that the HDP algorithm is not significantly affected by the need for clipping.

Fig. \ref{fig:resultsLLSmallerDeltaT} shows that the need for clipping is not made arbitrarily small by just using a smaller $\Delta t$ value.

\begin{figure}[ht]
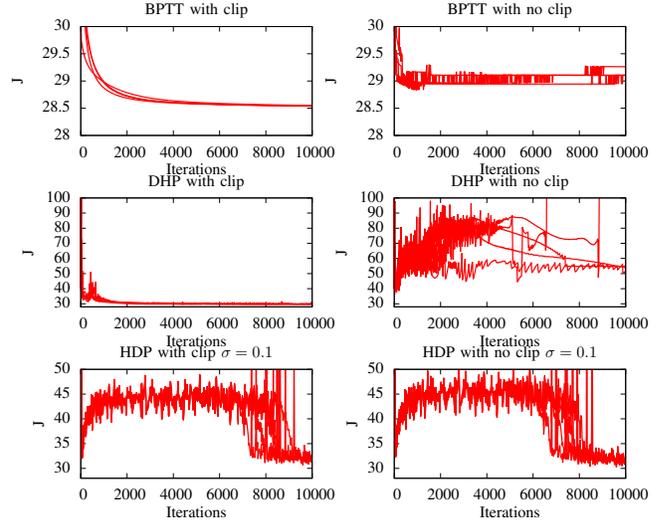

  \begin{center}
    \begin{tabular}{cc}
        \resizebox{35mm}{!}{\input{resultsLL/plotLL_BPTT_clipb}} &
        \ \ \resizebox{35mm}{!}{\input{resultsLL/plotLL_BPTT_noclipb}}\\ 
        \resizebox{35mm}{!}{\input{resultsLL/plotLL_DHP_clipb}} &
        \ \ \resizebox{35mm}{!}{\input{resultsLL/plotLL_DHP_noclipb}}\\
        \resizebox{35mm}{!}{\input{resultsLL/plotLL_HDP_clipb}} &
        \ \ \resizebox{35mm}{!}{\input{resultsLL/plotLL_HDP_noclipb}} 
    \end{tabular}
\caption {Vertical Lander solutions by BPTT, DHP and HDP using $\Delta t=1$.}
\label{fig:LL_performance}
  \end{center}
\end{figure}

\begin{figure}[ht]
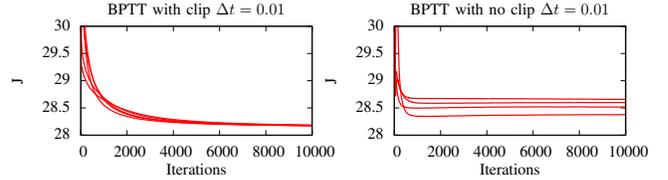

  \begin{center}
    \begin{tabular}{cc}
        \resizebox{35mm}{!}{\input{resultsLL/plotLLsmallDeltaT_BPTT_clipb}} &
        \ \ \resizebox{35mm}{!}{\input{resultsLL/plotLLsmallDeltaT_BPTT_noclipb}}
    \end{tabular}
\caption {Vertical Lander with $\Delta t=0.01$.}
\label{fig:resultsLLSmallerDeltaT}
  \end{center}
\end{figure}

\begin{table}
\caption {Terminal Boundary Planes used in vertical-lander experiment.  The state vector used here is $\vecx=(h,v,u)^T$.}
\begin{center}
  \begin{tabular}{ l  | l | l }
 Termination  & Position vector&  Normal Vector \\
 Condition Breached & of Plane, $\planeConstantP^T$ &  to Plane, $\vecn^T$ \\
\vspace{-0.1in}&&\\ 
\hline 
$h \le0$ (hits ground) & $\begin{smallmatrix}(0,0,0)\end{smallmatrix}$ & $\begin{smallmatrix}(1,0,0)\end{smallmatrix}$ \\
$u\le0$ (no fuel) &$\begin{smallmatrix}(0,0,0)\end{smallmatrix}$ & $\begin{smallmatrix}(0,0,1)\end{smallmatrix}$ 
  \end{tabular}
\end{center}
\label{table:terminationBoundaryPlanesLL}
\end{table}

\subsection{Cart Pole Experiment} \label{sec:cartPoleExpt}

\newcommand{\vecX}{{x}}
\newcommand{\vecF}{{\cartPoleForce}}

We investigated the effects of clipping in the well known cart-pole benchmark problem described in Fig. \ref{fig:cartPole}.  We considered the version of this problem used by \cite{barto83neuronlike}, where the total trajectory cost is a function of the duration that the pole could be balanced for.  Clearly, unless clipping is used properly, the duration will be an integer number of time steps, and since this is not smooth and differentiable, it will cause problems (become impossible) for DHP and BPTT.  Hence traditionally when DHP or BPTT are used for the cart-pole problem, a different cost function would be used, one that is differentiable and proportional to the deviation from the balanced position (e.g. see \cite{LendarisPendulumDHP}).  However in this section we show that by using clipping, DHP and BPTT can be successful with the duration-based reward.  Since it is not possible to do this without clipping, we assume this is the first published version of this solution by DHP/BPTT.

 \begin{figure} 
   \centering \scalebox{0.6}{
   \def\svgwidth{\columnwidth} 
   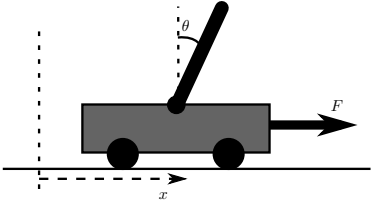}
  \caption{Cart-pole benchmark problem.  A pole with a pivot at its base is balancing on a cart. The objective is to apply a changing horizontal force $\cartPoleForce$ to the cart which will move the cart backwards and forwards so as to balance the pole vertically.  State variables are pole angle, $\theta$, and cart position, $\vecX$, plus their derivatives with respect to time, $\dot{\theta}$ and $\dot{\vecX}$.}
\label{fig:cartPole}
 \end{figure}

The equation of motion for the frictionless cart-pole system (\cite{barto83neuronlike,rflorianCorrectEquationsCartPole,LendarisPendulumDHP}) is: 
\begin{align}
 \ddot{\theta}&=\frac{g\sin{\theta}-\cos{\theta}\left[\frac{\cartPoleForce+ml{\dot{\theta}}^2\sin{\theta}}{m_c+m}\right]}
{l\left[\frac{4}{3}-\frac{m\cos^2{\theta}}{m_c+m}\right]} \\
\ddot{\vecX}&=\frac{\cartPoleForce+ml\left[\dot{\theta}^2\sin{\theta}-\ddot{\theta}\cos{\theta}\right]}{m_c+m}
\end{align}
where graviational acceleration, $g=9.8 ms^{-2}$; cart's mass, $m_c=1kg$; pole's mass, $m=0.1kg$; half pole length, $l=0.5m$; $\cartPoleForce \in [-10,10]$ is the force applied to the cart, in Newtons; and the pole angle, $\theta$, is measured in radians.  The motion was integrated using the Euler method with a time constant $\Delta t=0.02$, which, for a state vector $\vecx\equiv(\vecX, \theta, \dot{\vecX},  \dot{\theta})^T$, gives a model function $\modelFull{}=\vecx+(\dot{\vecX}, \dot{\theta}, \ddot{\vecX},  \ddot{\theta})^T\Delta t$.  

\begin{table}
\caption {Terminal Boundary Planes used in cart-pole experiment.  The state vector used here is $\vecx=(x,\dot{x},\theta, \dot{\theta},t)^T$.}
\begin{center}
  \begin{tabular}{ l  | l | l }
 Termination  & Position vector&  Normal Vector \\
 Condition Breached & of Plane, $\planeConstantP^T$ &  to Plane, $\vecn^T$ \\
\vspace{-0.1in}&&\\ 
\hline 
$\theta \ge \pi/15$ & $\begin{smallmatrix}(0,0,\pi/15,0,0)\end{smallmatrix}$ & $\begin{smallmatrix}(0,0,-1,0,0)\end{smallmatrix}$ \\
$\theta \le -\pi/15$ & $\begin{smallmatrix}(0,0,-\pi/15,0,0)\end{smallmatrix}$ & $\begin{smallmatrix}(0,0,1,0,0)\end{smallmatrix}$ \\
$x \ge 2.4$ & $\begin{smallmatrix}(2.4,0,0,0,0)\end{smallmatrix}$ & $\begin{smallmatrix}(-1,0,0,0,0)\end{smallmatrix}$ \\
$x \le -2.4$ & $\begin{smallmatrix}(-2.4,0,0,0,0)\end{smallmatrix}$ & $\begin{smallmatrix}(1,0,0,0,0)\end{smallmatrix}$ \\
$t \ge 300$ & $\begin{smallmatrix}(0,0,0,0,0)\end{smallmatrix}$ & $\begin{smallmatrix}(0,0,0,0,1)\end{smallmatrix}$ 
  \end{tabular}
\end{center}
\label{table:terminationBoundaryPlanesCartpole}
\end{table}

The pole motion continues until it reaches a terminal state or until the pole is successfully balanced for $300$ time steps, i.e. 6 seconds of real time.  Terminal states ($\TerminalStateSet$) are defined to be any state with $|\vecX|\ge2.4$, or  $|\theta|\ge\frac{\pi}{15}$ (i.e. 12 degrees), or $t\ge 300$.  Termination plane constants are given in Table \ref{table:terminationBoundaryPlanesCartpole}.

The duration-based cost function of \cite{barto83neuronlike} is defined as 
\begin{align}
 \reward(\vecx, \actiona)=&\begin{cases}
1 & \text{if $\vecx \in \TerminalStateSet$ and $t<300$} \\
0 & \text{if $\vecx \notin \TerminalStateSet$ or $t=300$}
\end{cases} \label{eqn:cartPoleCostFunction}
\end{align}
and when this is combined with a discount factor $\df<1$ gives a total trajectory cost of $\RpiFull{0}\equiv \df^{(\finalTimestep)}$, where $\finalTimestep$ is the trajectory duration.  Since this function decreases with $\finalTimestep$, mimimising it will increase $\finalTimestep$, i.e. lead to successful pole balancing. 

We tested the two algorithms BPTT and DHP on this problem with a discount factor $\gamma=0.97$.  To ensure the state vector was suitably scaled for input to the both MLPs, we used rescaled state vectors $\vecx'$ defined by $\vecx'=(0.16\vecX, 15\theta/\pi, \dot{\vecX},  4\dot{\theta})^T$, with $\theta$ in radians, throughout the implementation.  As noted by \cite{LendarisPendulumDHP}, choosing an appropriate state-space scaling can be critical to successful convergence of actor-critic architectures in the cart-pole problem.  The output of the action network, $y$, was multiplied by 10 to give the control force $\cartPoleForce=\PolicyFunctionFull{}\equiv10y$.  The learning rates for the algorithms that we used were: BPTT ($\learningRateActor=0.1$); (DHP: $\learningRateActor=0.01$,$\learningRateActor=0.0001$).  The DHP critic used a final-layer activation-function slope of 0.1.

Learning took place on five trajectories simultaneously, with fixed start points randomly chosen from the region $|\vecX|<2.4$, $|\theta|<\frac{\pi}{15}$, $\dot{\vecX}=0$, $\dot{\theta}=0$.  This is similar to the starting conditions used by \cite{barto83neuronlike}. The exact derivatives of the model and cost functions were made available to the algorithms.  

The performance of the two algorithms, both with and without clipping, are shown in Fig. \ref{fig:CP_performance}.  Each graph shows how the average balancing duration over all five trajectories, versus the training iteration.  Each graph shows an ensemble of five different curves, with each curve represents a training run from a different random weight initialisation.   

\begin{figure}[ht]
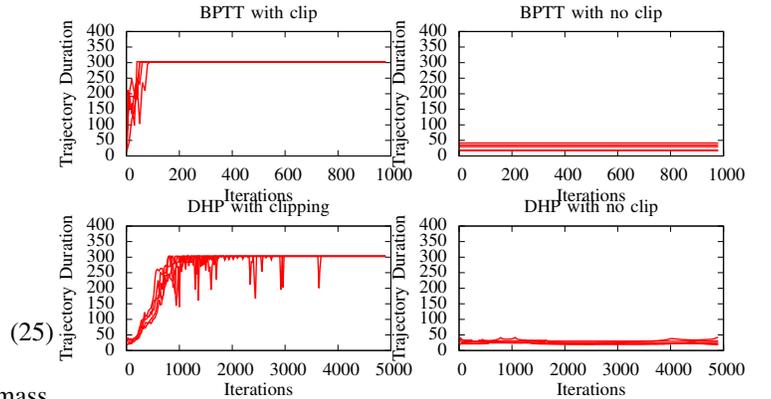

  \begin{center}
    \begin{tabular}{cc}
        \resizebox{40mm}{!}{\input{resultsCP/plotCP_BPTT_clipb}} &
        \resizebox{40mm}{!}{\input{resultsCP/plotCP_BPTT_noclipb}}\\ 
        \resizebox{40mm}{!}{\input{resultsCP/plotCP_DHP_clipb}} &
        \resizebox{40mm}{!}{\input{resultsCP/plotCP_DHP_noclipb}} 
    \end{tabular}
\caption {Cart-pole solutions by BPTT and DHP.}
\label{fig:CP_performance}
  \end{center}
\end{figure}

The results show that using clipping correctly enables both the DHP and BPTT algorithms to solve this problem consistently, and without clipping it is impossible for both algorithms.

\section{Conclusions} \label{sec:conclusions}

The problem of clipping for ADP/RL and neurocontrol algorithms has been demonstrated and motivated.  Without clipping, algorithms which rely on the derivatives of the model and cost functions can fail to work.  The solution is to apply clipping, and then to correctly differentiate the model and cost functions in the final time step.  This solution has been given in the form of the equations, plus in the form of clear pseudocode for the two major affected ADP algorithms: DHP and BPTT.

Two neural network experiments have confirmed the importance of applying clipping correctly.  These included a cart-pole experiment, where clipping was found to be essential, and in a vertical-lander experiment, where clipping produced a significant improvement of performance.  

The situations in which clipping are needed have been made clear, and those situation where it can be ignored have also been specified.

\bibliographystyle{IEEEtran}

\end{document}